# Evidential Reasoning in a Network Usage Prediction Testbed
(work in progress)

Ronald P. Loui*
Rockwell Science Center Palo Alto
444 High Street
Palo Alto, CA 94301
loui@csli.stanford.edu

Abstract.

This paper reports on empirical work aimed at comparing evidential reasoning techniques. While there is prima facie evidence for some conclusions, this is work in progress; the present focus is methodology, with the goal that subsequent results be meaningful.

The domain is a network of UNIX™ cycle servers, and the task is to predict properties of the state of the network from partial descriptions of the state. Actual data from the network are taken and used for blindfold testing in a betting game that allows abstention.

The focal technique has been Kyburg's method for reasoning with data of varying relevance to a particular query, though the aim is to be able to compare various uncertainty calculi, correctly applied.

The conclusions are not novel, but are instructive.

    1. All of the calculi performed better than human subjects, so unbiased access to sample experience is apparently of value.

    2. Performance depends on metric: (a) when trials are repeated, *net = gains - losses* favors methods that place many bets, if the probability of placing a correct bet is sufficiently high; that is, it favors point-valued formalisms; (b) *yield = gains/(gains+losses)* favors methods that bet only when sure to bet correctly; that is, it favors interval-valued formalisms.

    3. Among the calculi, there were no clear winners or losers.

Methods are identified for eliminating the bias of the *net* as a performance criterion and for separating the calculi effectively, in both cases by posting odds for the betting game in the appropriate way.

1. Testbed.

States of a UNIX™ cycle server network were described by asserting those sentences that hold in the state. For example,

    (AND (weekend) (in-use 'castor) (logged-on 'marsh) (on 'cox 'antares))

described a state in which the machine "castor" was in use, the user "marsh" was logged on, and the user "cox" was logged onto the machine "antares", some time during the weekend.

All such descriptions were unified with deductive rules and forward chained, so that the above description was augmented by

    (NOT (weekday)),    (logged-on 'cox),    (in-use 'antares),

and of course, the tautology,

    (always-true).

*this work was supported at the University of Rochester by a basic grant from the U.S. Army Signals Warfare Center and at Stanford University by a Sloan Cognitive Science Postdoctoral Fellowship. Author's current address: Dept. of Computer Science, Washington University, St. Louis, MO 63130. UNIX™ is a trademark of AT&T. The author thanks Jack Breese, Glenn Shafer, and an anonymous referee for comments, Michael McInerny, Bill Bolosky, Paul Cooper, Chris Brown, Cesar Quiroz, Dave Coombs, Joel Lachter, Leo Hartman, Eliz Hinkelman, and Josh Tenenberg for completing lengthy surveys, and the entire UR CS Dept. for sacrificing so many cycles.

The language allowed reference to such properties as machine load, number of users, kind of user, that is, whether a backup was in progress or there was a UNIX™-to-UNIX™-communication-protocal (uucp) user, number of ports, use of dial-in ports, and so forth. They could be combined with any logical connective, e.g.

    (OR (logged-on 'marsh) (logged-on 'cox))

and with some quantifiers. All quantities were discretized, so that the property

    (very-very-many-network-users)

was used instead of (> (# network-users) 20), for instance.

Given a state described partially, as above, what was at issue was the degree of belief in a sentence such as

    (logged-on 'jackson),

that is, the extent to which "marsh" and "cox" being so logged on a weekend provided evidence for "jackson" being logged on.

Several hundred snapshots of the network were taken at random times, that is, states of the network were observed. Each snapshot was associated with a time, and was roughly the result of concatenating the results of the "rwho", the "ruptime", and the "date" commands under UNIX™:

Some of these data were provided to a program as sampling data. Typically, sixty snapshots were provided. If sampling were done indeed randomly, we would expect that one-seventh of these, or about nine, would be during the weekend, and of those hypothetical nine weekend snapshots, perhaps two were when "marsh" was logged on, and three when "cox" was logged on. Perhaps there was even one in which both were logged on, and so was "jackson". Sometimes the record of available data was limited to twenty snapshots instead of sixty.

For each query, such as the probability of (logged-on 'jackson) given (AND (logged-on 'marsh) . . .), the snapshots were summarized as samples of varying relevance:

    (s% (  (AND (logged-on 'cox) (logged-on 'marsh))  (logged-on 'jackson) ) (4 1) )
    (s% (  (weekend) (logged-on 'jackson) ) (3 1) )
    (s% (  (AND (in-use 'castor) (logged-on 'marsh)) (logged-on 'jackson) ) (2 2) )
    (s% (  (always-true) (logged-on 'jackson) ) (20 4) )

The first statement, for instance, reports that in the record of snapshots, there were four times when both "cox" and "marsh" were logged-on, and in one of those four, "jackson" was logged-on too. In the general case, which includes twenty snapshots, "jackson" was logged-on four times. Note that in the record, there was no snapshot that captured all of (AND (weekend) (logged-on 'marsh) (on 'cox 'antares) (in-use 'castor)), nor even a snapshot in which (AND (weekend) (logged-on 'marsh) (logged-on 'cox)) held. So there are no samples from these classes.

    2. Betting.

The following process was repeated an arbitrary number of times.

A test snapshot was selected from those that did not constitute program data. Arbitrarily chosen properties were determined to hold of the state of the network captured by the snapshot, and an arbitrary number of them was announced. Typically three, eight, or sixteen properties were announced. Matching the discussion above, suppose the target property was (logged-on 'jackson) and four properties were announced: (weekend), (in-use 'castor), (logged-on 'marsh), and (on 'cox 'antares).

A lottery size, say 10, and payoff ratio, say .1, were chosen, arbitrarily, so that the following choice could be offered:

258

1. offer a pot of 10 for an ante of 1,
2. place an ante of 1 for a pot of 10, or
3. abstain.

If Belief( (logged-on 'jackson) given (AND (weekend) . . .) ) was less than .1, option 1 was the rational choice. If it exceeded .1, then option 2 was the rational choice. If belief were an interval that included .1, say Belief( (logged-on 'jackson) given (AND (weekend) . . .) ) = [.05, .3], then ostensibly, the rational choice was to abstain. Each program, or test subject, committed to one of these choices for each query.

The programs were forced to play the rational strategy for a single choice, without knowledge that the process was being repeated, that is, without knowledge that there would be more questions. The human test subjects were not prohibited from using the knowledge that the process was being repeated to alter their betting strategies, but were advised what would be the rational stragey for an unrepeated process. Any combination of choices was permitted, though some combinations required a larger financial base. For example, the strategy that constantly offers pots requires more intial resource than the one that constantly abstains. There was no sense in which any agent went bankrupt from too much loss.

The actual snapshot was consulted to determine whether the target property held. Payoffs were awarded appropriately.

This process was repeated under various conditions for various kinds of queries. Relative success was usually stable after some 50 repetitions, though some of the runs reported here repeat the querying over 1000 times under fixed conditions.

### 3. Uncertainty Calculi.

The original idea was to test Kyburg's method for determining probabilities from frequency information of varying relevance. It was to be compared against human performance. This paper reports those results.

This testbed, however, is natural for testing one uncertainty calculus against another. This paper also reports intial considerations for such testing. *These data must be taken as preliminary, as the central conclusions of this paper are methodological at this stage of the research.*

The non-Kyburgian methods used here are based on approaches that have been discussed in the uncertainty literature, but are not necessarily applied appropriately. For instance, the Dempsterian rule of combination is applied to intervals, rather than beginning with belief functions constructed from sampling data, as Shafer suggests [Shafer76]. The method of similarity [Salzberg86][Stanfill and Waltz87] is applied with an arbitrary similarity metric, with no attempt to optimize it for this domain. Subsequent research will apply these methods more appropriately. No prior probability information was used.

**Kyburg.** Kyburg's procedures [Kyburg74,82] start with frequency data in the form of intervals, for example

(% ( (weekend) (logged-on 'jackson) ) (.2 .4) ) ,

so sampling data must be converted into frequencies. This is done by approximating narrowest binomial confidence intervals at some level of confidence (usually .9). Some of these intervals can be combined on purely set-theoretic grounds. For example,

(% ( (in-use 'castor) (logged-on 'jackson) ) (.3 .5) )

can be combined with the interval above to yield

(% ( (XP (weekend) (in-use 'castor)) (X (logged-on 'jackson) (logged-on 'jackson)) ) (g(.2, .3) g(.4, .5)) )

where (XP (a) (b)) is

{<x, y> : (a) holds in x, (b) holds in y, and (logged-on 'jackson) holds in x just in case it holds in y}

259

where (X (a) (b)) is just a cross product: {<x, y> : (a) holds in x and (b) holds in y}, and g(p, q) = pq/(1 - p - q + 2pq). These constructions are discussed in [Loui86].

Of these intervals, Kyburg considers intervals that come from more specific sampling classes than intervals with which they disagree (intervals disagree if one does not nest in the other). The narrowest of such intervals under consideration is the probability.

Of the three intervals above, neither of the first two are considered because they disagree, and neither comes from a more specific class. The third disagrees with neither of the other two, so would be considered, if there were no other intervals.

**Loui.** Some reject Kyburg's rule for determining when intervals disagree. The intervals (.06 .63) and (.08 .73) disagree because neither is a sub-interval of the other. But neither would dispute the interval (.06 .73), and perhaps this could be taken as the probability. Kyburg's system was altered to allow this [Loui87]. This method for determining probabilities appears in the empirical work, though in not all of the runs, since it demanded considerably more computation.

**Naive Average.** Another method used in the empirical work takes the average of the maximum likelihood estimators for each sample, and this (point) value is taken to be the degree of belief. For n samples of sizes $s_i$ with results $r_i$ out of $s_i$, the belief would be

$$(1/n) \sum_{\text{all samples}} r_i / s_i .$$

With the four samples above, this would be [(1/3) + (1/4) + (2/2) + (4/20)]/4.

**Maximal Average.** Prior to averaging, estimators from sample classes are discarded if there is an estimator from a subclass. So a sample of snapshots in which (AND (in-use 'castor) (logged-on 'marsh)) held would supplant a sample in which just (in-use 'castor) held. Of the four samples above, only

(s% ( (always-true) (logged-on 'jackson) ) (20 4) )

is supplanted. It reports sampling from a superclass of (at least) one of the other classes reported, for example, (weekend), which is really (AND (weekend) (NOT weekday) (always-true)). So the (point) value would be

$$(1/[\# \text{ of maximal classes}]) \sum_{\text{samples from maximal classes}} r_i / s_i$$

or [(1/3) + (1/4) + (2/2)]/3. Note that non-maximal sets are discarded irrespective of the size of the sample from its subset, and irrespective of whether they disagree about the estimate.

**Similarity.** This method weighs each maximum likelihood estimator by the extent to which the sampled class matches the properties that are given. Also, samples contribute to the sum in direct proportion to their size. This is essentially the method used by Salzberg [Salzberg86] to predict horse race winners and by Stanfill and Waltz [Stanfill and Waltz87] to project pronunciations of words from a small dictionary.

There are many possible weighting schemes. The weights used here were linear in the number of common properties. If #common(set, given) is the number of given properties that hold of members of the set, for example

#common( (AND (always-true) (in-use 'castor)), (AND (weekend) (on 'cox 'antares) . . .) ) = 2,

then belief yielded by this method is

$$\sum_{\text{all classes}} \#\text{common}(\text{class}_i, \text{given})\, r_i / \sum_{\text{all classes}} \#\text{common}(\text{class}_i, \text{given})\, s_i$$

Above, including the (always-true) property, the (4 1) sample shares 3 properties with the given properties, the (3 1) sample shares 2, the (2 2) sample shares 3, and the (20 4) sample shares 1. So the degree of belief in (logged-on 'jackson) is

[(3)(1) + (2)(1) + (3)(2) + (1)(4)] divided by

$[(3)(4) + (2)(3) + (3)(2) + (1)(20)]$ .

**Naive Dempster.** With the intervals above, simple mass functions can be constructed. The frequency statement

(% (    (AND (logged-on 'marsh) (logged-on 'cox)) (logged-on 'jackson) ) (.06 .63) )

is converted to the mass function

$m_{\text{(AND (logged-on 'marsh) (logged-on 'cox))}}$ ( (logged-on 'jackson) ) = .06
$m_{\text{(AND (logged-on 'marsh) (logged-on 'cox))}}$ ( (NOT (logged-on 'jackson)) ) = .37
$m_{\text{(AND (logged-on 'marsh) (logged-on 'cox))}}$ ( (OR (logged-on 'jackson) (NOT (logged-on 'jackson))) ) = .57 .

This imputes a belief function that is combined with other belief functions constructed from other intervals. An alternative way of constructing belief functions would start with the sample data directly, as Shafer suggests [Shafer76]. Note that the belief functions constructed from

(% (    (logged-on 'cox) (logged-on 'jackson) ) (.08 .73) ) and from
(% (    (AND (logged-on 'marsh) (logged-on 'cox)) (logged-on 'jackson) ) (.06 .63) )

would be combined by Dempsterian combination under this method, though they do not represent clearly distinct evidential sources.

This method tended to yield narrow intervals, reflected in the fact that it did not abstain much.

**Maximal Dempster.** Like maximal average, sources are discarded if they report on statistics about a class, and there is knowledge of statistics about some subclass. So of the above two frequency statements, since (logged-on 'cox) is not as specific as (AND (logged-on 'marsh) (logged-on 'cox)), a belief function is constructed only for the latter, and combined with other maximal sources of statistical information.

This method tended to yield the widest intervals of all the methods considered here; it abstained most often.

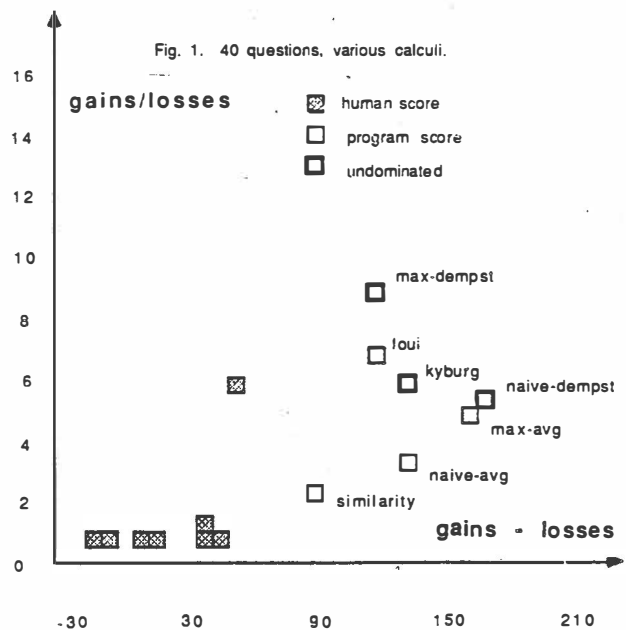
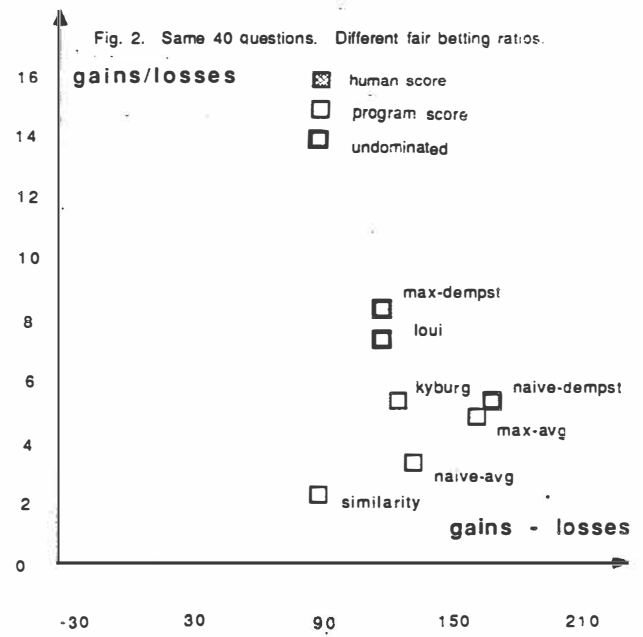

Fig. 1. 40 questions, various calculi.

Fig. 2. Same 40 questions. Different fair betting ratios.

## 4. Some Preliminary Data.

A questionnaire with 40 choices was completed by some users of the network with varying knowledge of the patterns of its use, and the questions were also answered by each of the techniques described. Human respondents did not have access to summarized sample information, but presumably had their own prior

261

experience. It would be interesting to test the opposite situation, wherein agents respond only to the same syntactic information provided to the programs, with no additonal exposure to the domain.

Table 1 summarizes the results, sorted by net, which was gains less losses. An alternative performance criterion wasy yield, which was gains/(gains+losses), or gains/losses. Figure 1 plots performance using both metrics.

Table 1. 40 questions sorted by net. %max is the per cent of the perfect score. %rel is the per cent of the highest achieved score. #absts reports the number of times the subject chose to abstain. data indicates the number of data points each program accessed, or else the familiarity of the subject with the usage patterns on the UNIX™ network.

| subject | data | net | %max | %rel | gains | losses | g/l | yield | #absts |
| --- | --- | --- | --- | --- | --- | --- | --- | --- | --- |
| naive dempster | 60 | 172.5 | 68 | 100 | 213 | 41.5 | 5.1 | .84 | 0 |
| maximal average | 60 | 160 | 63 | 93 | 207.5 | 47.5 | 4.4 | .82 | 0 |
| kyburg | 60 | 136.5 | 54 | 79 | 168.5 | 32 | 5.3 | .84 | 6 |
| naive average | 60 | 136.5 | 54 | 79 | 195.5 | 59 | 3.1 | .76 | 0 |
| maximal dempster | 60 | 126.5 | 50 | 73 | 142 | 15.5 | 9.2 | .90 | 13 |
| loui | 60 | 123 | 48 | 71 | 146 | 23 | 6.3 | .86 | 12 |
| similarity | 60 | 92.5 | 36 | 53 | 173.5 | 81 | 2.1 | .68 | 0 |
| ai student | good | 57 | 22 | 33 | 69 | 12 | 5.8 | .85 | 24 |
| systems student | good | 44.5 | 16 | 25 | 149 | 105.5 | 1.4 | .58 | 0 |
| author | good | 37.5 | 14 | 21 | 99 | 61.5 | 1.6 | .62 | 17 |
| vision student | fair | 32 | 12 | 18 | 89.5 | 57.5 | 1.6 | .62 | 22 |
| vision faculty | fair | 19.5 | 7 | 11 | 72 | 52.5 | 1.4 | .58 | 16 |
| systems student | good | 11 | 4 | 6 | 58 | 47 | 1.2 | .55 | 19 |
| ai student | fair | -2 | - | - | 77 | 79 | 1.0 | .50 | 16 |
| psych student | poor | -7 | - | - | 117.5 | 124.5 | 0.9 | .50 | 2 |

It was clear from the data that the uncertainty calculi, with access to unbiased sampling data, were outperforming the human subjects. In aggregate net, every system did better than every human subject. In yield, (which shares the same ordering as gains/losses), only one human subject did as well, though his score was quite high.

Note that subjects performed poorly regardless of strategy: some tended to choose lotteries of which they were most certain and abstained frequently; others guessed recklessly. Even when subjects thought they knew the answer, they did not, so their yields suffered despite careful selection of bets to place. Note also that there was some correlation between subjects' familarity with the system and with performance.

Subjects were observed to manifest all of the classic judgement biases [Tversky and Kahneman74]. For instance, the representativeness heuristic led most to overestimate the probability of (logged-on 'cheng) given there was a (backup-somewhere), since "cheng" was the user responsible for doing backups. But of course, backups ran much longer than cheng's logons.

It is not clear whether net or yield should be the performance criterion. Apparently, net favored agents who did not abstain much, while yield (or gains to losses) favored agents who were cautious about the choice of bets, so long as they did well on those bets that they did choose. *Had table 1 sorted by yield (or gains/losses), the ordering among the programs would have been largely reversed.*

Figures 2 and 3 show performance of the programs for different choices of payoff ratios (figure 2), and for different payoff ratios and different lottery values (figure 3). *Relative order among the programs is not robust for this small number of repetitions.*

Figure 4 shows the effect of limiting programs to 20 data points instead of 60. Sample sizes that may have been

        (s% (   (AND (logged-on 'marsh) (logged-on 'cox) ) (logged-on 'jackson) ) (4 1) )

may be reduced to

        (s% (   (AND (logged-on 'marsh) (logged-on 'cox) ) (logged-on 'jackson) ) (1 0) )

or disappear entirely. For methods that produce wide intervals, the gains/losses performance jumped because of more frequent abstaining. Diminished access to data had a detrimental effect on net, in general.

[Loui85] suggests altering boldness, or width of intervals, as the size of stakes is altered. Table 2 gives prima facie evidence that this had a positive effect, though much more investigation needs to be done.

Table 2. Result of altering confidence levels. Kyburg(.7) applies Kyburg's methods to intervals constructed from samples at .7 confidence. Kyburg(.9) applies to intervals at .9 confidence. Kyburg(.7, .9) alters the confidence level between .7 and .9 depending on lottery size: the larger the lottery, the lower the confidence level appropriate for analysis of decision. 40 queries.

| method | %rel | gains/losses |
|---|---|---|
| kyburg(.7, .9) | 54 | 5.3 |
| kyburg(.7) | 50 | 4.0 |
| kyburg(.9) | 47 | 4.8 |

Again, wider intervals (from higher confidence level) results in more abstatining, a lower net, and a higher yield, all other things being equal.

We also investigated the effect of varying the number of properties of each snapshot announced for each blindfold test. It did not matter much. We had hypothesized that the maximal versions of the various calculi would fare better than the naive versions as the number of given properties increased. Intuitively, it seemed that the disparity in relevance between samples from classes such as

(AND (weekend) (in-use 'castor) (on 'cox 'antares) (logged-on 'marsh) (on 'lata 'lesath) . . .) and (weekend)

would grow with increased number of properties. Clearly the more specific class should then bear more of the weight in determining the probability. Methods insensitive to such specificity would be confused. This was not observed in the data. Table 3 shows the result of 1650 queries.

Table 3. Different number of announced properties. The second column shows %rel for 8 properties as a percentage of %rel for 3 properties. The third column shows %rel for 16 properties as a percentage of %rel for 3 properties. Recall that %rel is the percentage of the method's net relative to the highest net among programs, for a particular run. With 16 given properties, assuming no forward chaining, there are potentially 2^16 sampling classes. Of course, individuation among these classes is limited by the number of total data points, which was 60. 1650 queries total.

| method | 3 properties %rel%of3 | 8 properties %rel%of3 | 16 properties %rel%of3 |
|---|---|---|---|
| kyburg | 100 | 102 | 108 |
| similarity | 100 | 84 | 101 |
| maximal average | 100 | 101 | 100 |
| naive dempster | 100 | 98 | 100 |
| naive average | 100 | 89 | 99 |
| maximal dempster | 100 | 97 | 91 |

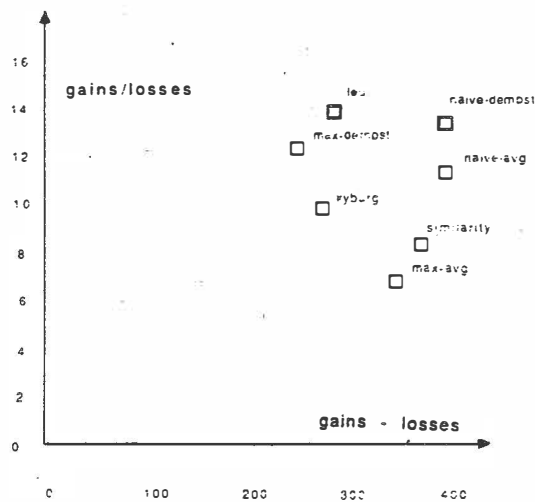

Fig. 3. Same 40 questions. Different fair betting ratios and lottery values

Only Kyburg's method fared consistently better with increased number of properties. In fact, the maximal version of Dempsterian combination fared successively more poorly, and the naive version seemed unaffected, with more numerous announced properties.

Although the relative performance of methods was fairly close for most runs, this should not lead us to complacency in our choice of uncertainty calculus, nor dissuade us from further investigation. The relative performance of these calculi can be made to differ markedly, when most of the choices are difficult. On a run of 40 queries, over half of the payoff ratios were such that all of the calculi agreed on the appropriate choice. It may have been obvious that the probability of (logged-on 'jackson) did not exceed .8, for instance. Table 4 shows the relative performance, in terms of net, on 750 queries when the payoff ratio was set by examining the belief of the programs.

Table 4. Relative net on difficult choices. In the first column, the payoff ratio equals the belief reported by the naive average method. In the second column, the payoff ratio equals the average of the beliefs of all the methods, where midpoints of intervals are used for interval-valued beliefs. 750 queries.

| method | odds set by naive average %rel | odds set by average of beliefs %rel |
|---|---|---|
| naive dempster | 100 | 100 |
| maximal average | 98 | 92 |
| kyburg | 77 | 85 |
| naive average | -- | 68 |
| similarity | -4 | 65 |
| maximal average | 33 | 5 |

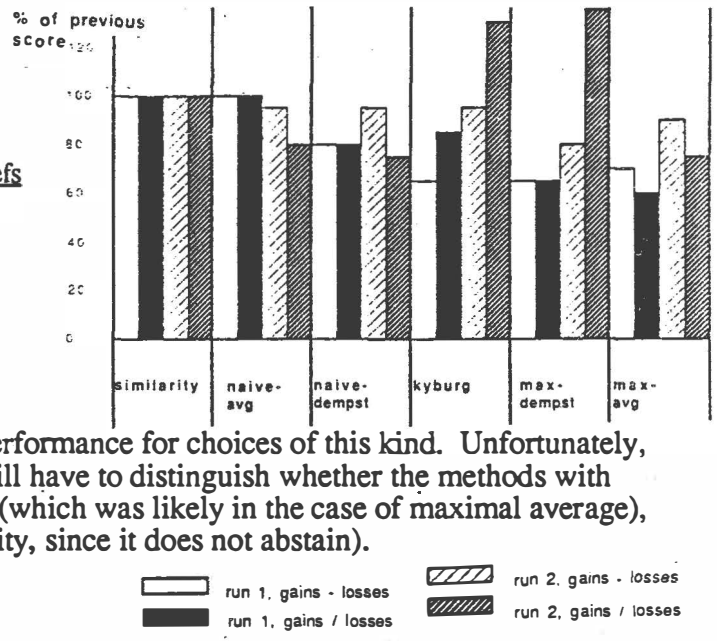

Fig. 4. Effect of using 20 data points instead of 60 from which to calculate probabilities. Note that for the wide-interval methods, the gains/losses score jumps because of more frequent abstaining.

Legend: run 1, gains - losses; run 1, gains / losses; run 2, gains - losses; run 2, gains / losses.

Apparently, there could be large differences in relative performance for choices of this kind. Unfortunately, yield data was not kept for these queries. Future work will have to distinguish whether the methods with low nets fared as they did because of frequent abstaining (which was likely in the case of maximal average), or just poor guessing (which must be the case for similarity, since it does not abstain).

## 6. Bias of Net for Repeated Choices.

The major problem encountered above was defining the performance criterion. Both net and yield are sensible, and the partial order under <net, yield> tuples is a natural compromise. It is natural to focus attention on the "pareto" frontier of undominated methods. In fact, one can study the convex hull that results by playing mixed strategies of method: this defines the frontier of <net, yield> points that we know how to achieve.

But at the moment there is a problem with the net as a performance criterion. It is biased in favor of those who guess often, who abstain very little. This is okay, if guesses are good. It makes sense to reward repeated good guessing, and to penalize methods which lack courage.

How bad can a guess be, yet still be worth making? The calculations of figures 5 and 6 suggest that guesses are worth making if the probability of guessing right exceeds 1/2. Suppose an interval-valued belief straddles the payoff ratio, for example, [.4, .52] straddles .5. Nevertheless, when an interval straddles the payoff ratio in this way, the probability is greater than .5 by epsilon that the correct choice is to offer the pot (that is, to act as if the belief was entirely below .5). Even for small epsilon, it can be shown under certain assumptions that the expected gain is positive if the bet is taken. This is bad news for those methods that abstain.

Perhaps this invites too much courage in betting. Under repeated trials, this expected gain is manifest. If p is the probability of guessing correctly, independently of whether the guess was to ante or to offer the pot, and the process is repeated $l_i$ are the lottery values, $i = 1, \ldots, k$, then

$$(\Sigma_{1..k}\, l_i\, )(p - .5)$$

is the expected net. The expected yield is always p. Prior to analysis of individual choices, it can be determined that the optimal strategy is to bet twice, instead of once, or no times, if two choices are offered (so long as betting is done with a greater than one-half chance of betting correctly). Having made the commitment to place all bets offered, even interval methods can then be applied to determine which bet to place: whether to ante or offer the pot, having already ruled out abstaining. If interval methods do not legislate a choice, it is still optimal to choose one or the other out of indifference.

This is not how interval-valued bet-placing strategies were implemented. So point-valued methods are favored to perform well in aggregate net because they implement the optimal multi-stage solution by nature. We are not interested in which method performs best on the multi-stage decision. We are interested in repeating the one-stage decision so that we obtain plausible long-run data about relative performance on one-stage bets. If we were interested in plausible data about relative performance on 40-stage bets, we would repeat play on the 40-stage game a few hundred times. Again, the optimal 4000-stage strategy might not be the same as the optimal 40-stage strategy, and methods that happen to implement the same strategy for both will be at an advantage.

Nevertheless, it makes sense to reward methods that actually place good bets.

One remedy is to set odds very close to the degrees of belief yielded by each method, as described in table 4. In this way, it is not a trivial accomplishment to guess correctly more than one-half the time. In fact, when payoff ratios are taken to be the average of the various degrees of belief, the number of correct guesses, if guesses were forced and abstaining disallowed, should equal the number of incorrect guesses.

## 7. Future Work.

This research began with an interest in showing that Kyburgian methods did what they were supposed to do: begin with raw, but limited sampling data of varying relevance, in a complex domain, and produce degrees of belief which are better in some sense than those that human experts would produce.

But the more interesting question is whether it was unbiased and indefatigable access to data that improved performance, or whether Kyburg's rules for processing the data were themselves responsible for success. Some preliminary study was made, with the following recommendations for future work.

In order to separate the methods effectively by relative performance, payoff ratios should be determined so that the probability of guessing correctly, p, is close to one-half. One way of doing this is taking the payoff ratio to be the average of the degrees of belief produced by the various methods. With p near one-hald, the bias inherent in the use of aggregate net is tempered. Methods should still be partially ordered by <net, yield> dominance. Undominated methods deserve attention, especially the methods that define the convex hull of performance. If the primary goal is comparison of techniques, techniques should be applied properly. This includes constructing belief functions from data, instead of interval, trying different similarity metrics, and altering confidence levels according to sizes of lotteries.

The prima facie conclusion is that the rules themselves are not responsible for the observed success: super-human performance was obtained with any of a number of methods. Nevertheless, subsequent research ought to be able to separate the better among them.

## 8. Bibliography.

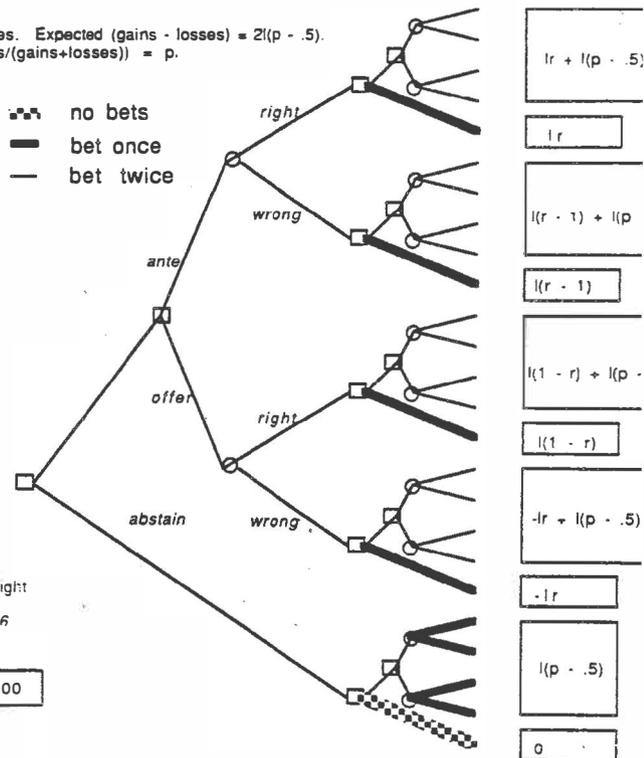

Fig. 6. Two choices. Expected (gains - losses) = 2l(p - .5). Expected (gains/(gains+losses)) = p.

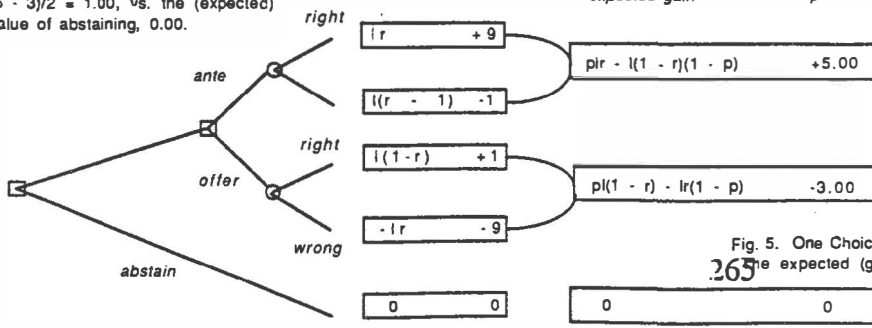

if anteing is as likely as offering, then the expected value of betting is (5 - 3)/2 = 1.00, vs. the (expected) value of abstaining, 0.00.

Fig. 5. One Choice. The expected (gains - losses) = l(p - .5). The expected (gains/(gains+losses)) = p.

265